\documentclass{2018-Shai}

\title{Learning the Geometric Mechanics of Robot Motion Using Gaussian Mixtures}
\author{Ruizhen Hu and Shai Revzen}
\journalname{Learning Geometric Mechanics with A-GBR}

\usepackage{moreverb,url}

\usepackage[colorlinks,bookmarksopen,bookmarksnumbered,citecolor=red,urlcolor=red]{hyperref}

\usepackage{amsmath, amssymb, bm}

\usepackage[utf8]{inputenc}
\usepackage{graphicx}
\graphicspath{{.}{./figs}{./figs/multipod/multipod_Leg6_wz}{./figs/robot}{./figs/stashed-PDFs}}
\usepackage{float}
\usepackage{subcaption}
\usepackage{xcolor}
\usepackage{hyperref}

\usepackage[]{natbib}

\newcommand{\T}{\mathsf{T}}

\newcommand{\group}{\mathfrak{G}}
\newcommand{\SE}{\mathsf{SE}}

\newcommand\BibTeX{{\rmfamily B\kern-.05em \textsc{i\kern-.025em b}\kern-.08em
T\kern-.1667em\lower.7ex\hbox{E}\kern-.125emX}}

\begin{document}

\footnote{Corresponding author \href{mailto:shrevzen@umich.edu}{shrevzen@umich.edu}}

\maketitle

\tableofcontents

\section*{Abstract}
Data-driven models of robot motion constructed using principles from Geometric Mechanics have been shown to produce useful predictions of robot motion for a variety of robots.
For robots with a useful number of DoF, these geometric mechanics models can only be constructed in the neighborhood of a gait.
Here we show how Gaussian Mixture Regression (GMR) can be used as a form of manifold learning that learns the structure of the Geometric Mechanics ``motility map''\footnote{In the geometric mechanics literature, what we refer to as the ``motility map'' here is sometimes called the ``(negative of the) local connection''. 
This terminology reflects its mathematical provenance from the fiber bundle literature, and not its application. 
We have therefore chosen a more readable name for use here.} and demonstrate: [i] a sizable improvement in prediction quality when compared to the previously published methods; [ii] a method that can be applied to any motion dataset and not only periodic gait data; [iii] a way to pre-process the dataset to facilitate extrapolation in places where the motility map is known to be linear.
Our results can be applied anywhere a data-driven geometric motion model might be useful.

\section{Introduction}
Geometric Mechanics describes the decomposition of (Lagrangian) dynamics under a symmetry into ``body frame'' and ``shape'' coordinates.
In environments where friction or viscosity annihilates body momentum sufficiently quickly, the consequence is a first-order equation of motion which is on its face quite unlike Newtonian dynamics -- it is ``geometric'' in the sense that only the shape of the motion matters, but the speed does not.

\subsection{A Brief Review of Classical Geometric Mechanics}

Consider a collection of interconnected bodies whose configuration $q$ is subject to forces $F(q,\dot q)$ that are equivariant under changes of position and orientation due to a Lie group $\group$, such as $\group=\mathsf{SE}(3)$.
The equivariance means that $F( g(q), ad_g(\dot q) ) = ad^*_g F(q, \dot q)$ i.e. for any group element $g \in \group$, changes of the coordinate frame transform forces to the forces that would have arisen from the transformed positions and velocities.
Typically this arises by having multiple body parts whose positions relative to the body frame are defined by its ``shape'', and having the forces arise through contact interactions of those body parts with an environment which is homogeneous (outside perhaps the influences of the body on its immediate environs).
From the mathematical perspective, ``shape'' is not the fundamental concept.
Rather, it is ``shape'' that arises from the equivalence of body configurations under the Lie group action, i.e. shape is the equivalence class $r := [q]_\group$.
The configuration $q$ thus partitions into a changing ``(body) shape'' $r$ and a moving body frame $g$, with an associated body velocity in the relevant Lie algebra, $v_b \in \mathcal{A}_\group$ (e.g. $\mathsf{se}(3)$ for $\SE(3)$).

The resulting dynamics lead to the ``reconstruction equation''
\begin{align}
	v_b = A(r) \dot r + \mathbb{I}^{-1}(r) p
\end{align}
for some momentum $p$ associated with the moving body frame, and expressed through a shape dependent ``frozen inertia tensor'' $\mathbb{I}(r)$.
For many systems, $p=0$, or $p \to 0$ so quickly that the $p = 0$ assumption is reasonable, leaving only $A(r)$ to govern the motion; such systems are called ``principally kinematic'', and the absence of $p$ from these equations lends ``geometric mechanics'' its name as ``geometric''.
The $A(r)$ term is known as the ``local connection'' or ``motility map'', and estimating it from data is the subject of this paper.

\subsection{Background}

In 1989 Shapere and Wilczek published a physics paper that arguably started the field of geometric mechanics, by showing that shape-change in micro-organisms produced motion that is startlingly different from Newtonian mechanics intuitions (\cite{shapere1989geometry}).
In the time since, the locomotion of many biological and robotic systems across diverse terrains has been shown to be principally kinematic.
Examples include locomotion on granular substrates like sand (\cite{astley2020surprising}), and multi-legged locomotion on solid ground (\cite{zhao2022walking}).
Morphologies studied include centipede-like robots, as in \cite{zhao2022walking, chong2023self} and snake robots, as in \cite{vaquero2024eels}.

 The complexity of body-environment interactions is hard to model in multi-contact or distributed contact situations.
According to \cite{astley2020surprising}, conventional approaches to multi-legged locomotion would require a combinatorial number of domains to model all contact states, and models of motion on sand would rely on having dynamical models for granular fluids which (so far) do not exist except in rare cases.
While some groups are trying direct modeling approaches (\cite{prasad2023geometric, wu2024modeling}), data-driven methods are increasingly employed for modeling and control of principally kinematic systems (\cite{deng2024adaptive, rieser2024geometric}).

Data-driven approaches are effective because the motility map is much easier to learn from data than more general cases of Newtonian dynamics.  
Once it is learned, various downstream tasks, such as control and gait optimization, can be efficiently performed as shown in \cite{hatton2011geometricMP, ramasamy2019geometry, yang2024towards}.

In previous work, our group has demonstrated the means to learn the motility map model in the neighborhood of a periodic motion (\cite{bittner2018geometrically}), and extended these learning methods to systems with non-zero momentum (\cite{kvalheim2019gait}), and under-actuated systems with passive degrees of freedom (\cite{bittner2022data}).
We have also shown that data-driven motility map models are highly predictive for multi-legged locomotion with or without slipping in both animals and robots as demonstrated by \cite{zhao2022walking}.
Follow-up work by Bittner also showed that these models could be updated with online learning (\cite{deng2024adaptive}).

However, data-driven motility map models remained either ``easy'' to identify near a periodic gait, or several related gaits (\cite{bittner2021optimizing}), or ``exponentially hard'' to identify, requiring an amount of data exponential in the number of DoF for an exhaustive model (\cite{dai2016geometric}) throughout the phase space.

\subsection{Contribution}

Our work here provides several contributions:
\begin{enumerate}
  \item We reformulate the problem of learning a motility map $A(r,\dot r)$ as a multi-valued manifold learning problem, allowing for non-linearities in $\dot r$ such as those predicted by \cite{kvalheim2019gait}, other non-linear structures in $r$ and $\dot r$, and hysteretic multi-valued motility maps to be represented.
  \item We provide a data transformation that allows a degree of linear extrapolation in the $\dot r$ direction to be controlled, capturing the fact that the manifold representing the motility map is often close to being a ruled surface with $A(r,\dot r)=A(r)\cdot \dot r$.
  \item We present an adaptation of well-known Gaussian Mixture Regression techniques to the machine learning problem of regressing a multi-valued function with a strong prior towards linearity in some parameters.
  \item We demonstrate that when applied to the datasets of \cite{zhao2020multi} and \cite{Zhao-2021-PhD} (available at \cite{BigAntData}, \cite{MultipodData} respectively), consisting of multi-legged robot data, this new approach noticeably improves the predictive ability relative to the prior state of the art.
\end{enumerate}

\section{Methods}

\subsection{Manifold Learning}

Manifold learning is the process of estimating the structure of a manifold from a set of noisy observations of points taken from that manifold, usually while the manifold is embedded in a larger space and the noise can offset the observations in that larger space. 
This form of learning operates on the assumption that high-dimensional data often lies on or near a lower-dimensional manifold embedded within the higher-dimensional space.
Since every smooth function defines and is uniquely described by a manifold which is its graph, methods of estimating manifolds applied to the input-output pairs of a function are closely related to methods of function regression. 

The most commonly familiar form of this relationship between estimating functions and fitting their graphs is the relationship between (Ordinary) Linear Least Squares Regression, which is a form of function estimation, and Total Least Squares (TLS) Regression which is a method for estimating the best fit hyperplane through the data -- a form of manifold regression restricting the class of manifolds to hyperplanes.
Because all $k$-dimensional manifolds have the property that at any point $p$ they are locally close to a $k$-dimensional affine subspace -- the manifold's tangent at $p$ -- one can approximate any manifold as a mixture of local TLS regressions estimating these tangent spaces.

From a practical standpoint, many functions modeling physical systems are applied to a continuous input -- either time, or some continuous function of time, and can be multi-valued when treated as a function of the current input.
Such multi-valued functions take on different values for the same input depending on the path used to reach it.
The most common example of these functions are hystereses, such as blacklash in gear mechanisms, diode hysteresis in electronics, and robot contact force hysteresis in contacts made through rubber-padded limbs.
To capture this kind of multi-valued behavior one may build a model $y = f(x,y_p)$ where the output $y$ is taken as a function of $x$ and a recent output $y_p$, usually arising from the previous time-step\footnote{This can also be thought of as a dynamical model $y_{n+1} = f(y_{n};x)$ where $x$ is a control input, constructed over dynamics where the control is ``very high gain'', allowing for only a few values of $y$ to be possible for any control $x$. We do not further explore this perspective here.}.
Using this approach, the model ``chooses'' a branch of the possible multi-valued outputs and remains on that branch.
If an observer estimating the true value of $y$ is used to inform the value of $y_p$ and $(x,y_p)$ is near a branching point where $y$ can bifurcate as $x$ is driven through it, the model will follow whichever branch of values is most consistent with the observed $y$ --- a highly useful property in practice.
Note that if we require the condition $f(x,y_p) \to y$ for all $y_p \to y$, the function $f$ will be continuous; this will come into play below.

Here we suggest that the motility map's graph be considered as such a multi-valued function, and we attempt to model it using Gaussian Mixture Regression (GMR).

\subsection{Gaussian Mixture Regression of Motility Maps}

In this section, we mathematically describe how a Gaussian Mixture Model (GMM-s) and the closely related technique of Gaussian Mixture Regression (GMR) are able to model a motility map.

Consider the following smooth manifold representing the graph of the motility map $A(r)$ in an ambient space $X$:
\begin{align}
	P &= \{(v_b, r, \dot{r}) \in X | v_b = A(r)\dot{r} \}&& X := \mathbb{R}^{N_b+2N_s},
\end{align}
where $N_b, N_s$ are the dimension of body space and shape space respectively. 
We assumed there exists some probability measure $d\mu_P(p)$ on $P$ representing the sampling process that generated our data in $P$, and some ``measurement error'' probability distribution $\text{Pr}(x | p)$ on the ambient space for each point $p\in P$.
This defines a measurement probability distribution on the ambient space
\begin{align}
  \Pr(x) = \int_P \Pr(x | p) d\mu_P(p)
\end{align} 

Our approach is to approximate the density $\Pr$ with a Gaussian Mixture Model (\cite{mclachlan2000finite}), and use the predictions obtained from these Gaussian components to provide a Gaussian Mixture Regression value for $A(r)\dot r$.
Following \cite{sung2004gaussian} we note that were we only predicting outputs from inputs (without additional recent inputs), this would be a direct application of the Nadaraya-Watson estimator (\cite{nadaraya1964estimating, watson1964smooth}), and it is established that Nadaraya-Wastson estimators are universal estimators for continuous functions .

\subsubsection{Gaussian Branching Regression}
\newcommand{\Aut}{\text{Aut}}
For completeness, we describe our approach from first principles, building up from the individual Gaussian components of the GMM modeling the graph.

Take a linear model $y = A x$, $x \in X$, and $y \in Y$, applied to a sample of $x$ points coming from a Gaussian distribution $N(\mu,\sigma)$.
The graph of this linear model is a point set in the direct sum vector space $W := X \oplus Y$.
We use the direct sum operator $\oplus$ to remind the reader that the vector space structure of the spaces $X$ and $Y$ extends naturally to $W$ as $(x \oplus y) + \alpha (v \oplus u) := (x +\alpha v) \oplus (y + \alpha u)$.
Additionally, any linear map $M$ from $W$ to itself, i.e. $M\in\Aut(W)$ has four parts: $M_{xx} \in\Aut(Y)$, $M_{yy}\in\Aut(Y)$, $M_{xy}\in L(Y,X)$ and $M_{yx} \in L(X,Y)$, such that $M (x \oplus y) := (M_{xx} x + M_{xy} y) \oplus (M_{yx} x + M_{yy} y)$ following the usual block structure of matrix operations.
When $M_{xy}$ and $M_{yx}$ are both zero, we write the block diagonal $M$ as $M_{xx} \oplus M_{yy}$, noting that in this case $(M_{xx} \oplus M_{yy}) (x \oplus y) = (M_{xx} x) \oplus (M_{yy} y)$.

Given the distribution of $x$ is Gaussian, the graph of points $(x,Ax)$ is itself a sample from a (degenerate) Gaussian $N( \mu \oplus A \mu, S)$, where $S$ is a rank $\dim X$ symmetric non-negative matrix whose $S_{xx}$ block is $\sigma$ and whose $S_{yy}$ block is $A \sigma A^\mathsf{T}$.
Assuming the measurements of $x$ (resp. $y$) are corrupted by (independent) Gaussian measurement noise with covariance $C_x$ (resp. $C_y$) the noisy data of the graph is still a sample from a Gaussian, now $N( \mu \oplus A \mu,~ S+(C_x \oplus C_y))$, and no longer degenerate. 

Assuming $\sigma$ is sufficiently larger than $C_x$ and $C_y$, the matrix $A$ can be recovered from $S$ by using Total Least Squares regression.
The TLS computation splits $W$ space along the eigenspaces of $S$, by decomposing $S = U^\mathsf{T} d U$ with $U$ orthogonal and $d$ a non-negative diagonal matrix with non-increasing elements along the diagonal.
The TLS estimate of the graph is the kernel of the $\dim Y$ last rows of $U$, i.e. the space $U_{yx} x + U_{yy} y = 0$, or 
\begin{align}
  y &:= A x = - U_{yy}^{-1} U_{yx} x. \label{eqn:TLS}
\end{align}
Given a Gaussian input distribution and Gaussian measurement noise for input and output we conclude that the measurements of the resulting graph follow a Gaussian distribution as well.
Furthermore, with sufficiently small measurement noise this graph Gaussian can be used to estimate the output for any input by treating the graph Gaussian's covariance as a linear constraint. 

To estimate a multi-valued function we combined the estimates provided by different components of the GMM of its graph based on their respective weights.
For classical GMR, the weights would be $w_i \Pr( x | N(\mu_i,\sigma_i) )$ -- the weight $w_i$ of the Gaussian component in the mixture, multiplied by the probability of the $x$ value for that Gaussian component.
Instead we used $w_i \Pr( (x,y_p) | N(\mu_i,\sigma_i) )$, taking into account the previous output $y_p$.
When multiple branches of $y$ values existed for an $x$, and were distinct with respect to the measurement noise, the Gaussians corresponding to $y$ values on branches other than that of $y_p$ had vanishingly small probabilities and thus little to no influence on the interpolation results.
The result is what we call a ``Gaussian Branching Regressor (GBR)'' which is defined on input of the form $(x,y_p)$ instead of $x$, and sensibly selects a branch and remains within it when computing a path integral of e.g. the motility map from data that seems to be multi-valued.

\subsubsection{An illustrative example}
\begin{figure}[h]
  \centering
  \includegraphics[width=.48\textwidth]{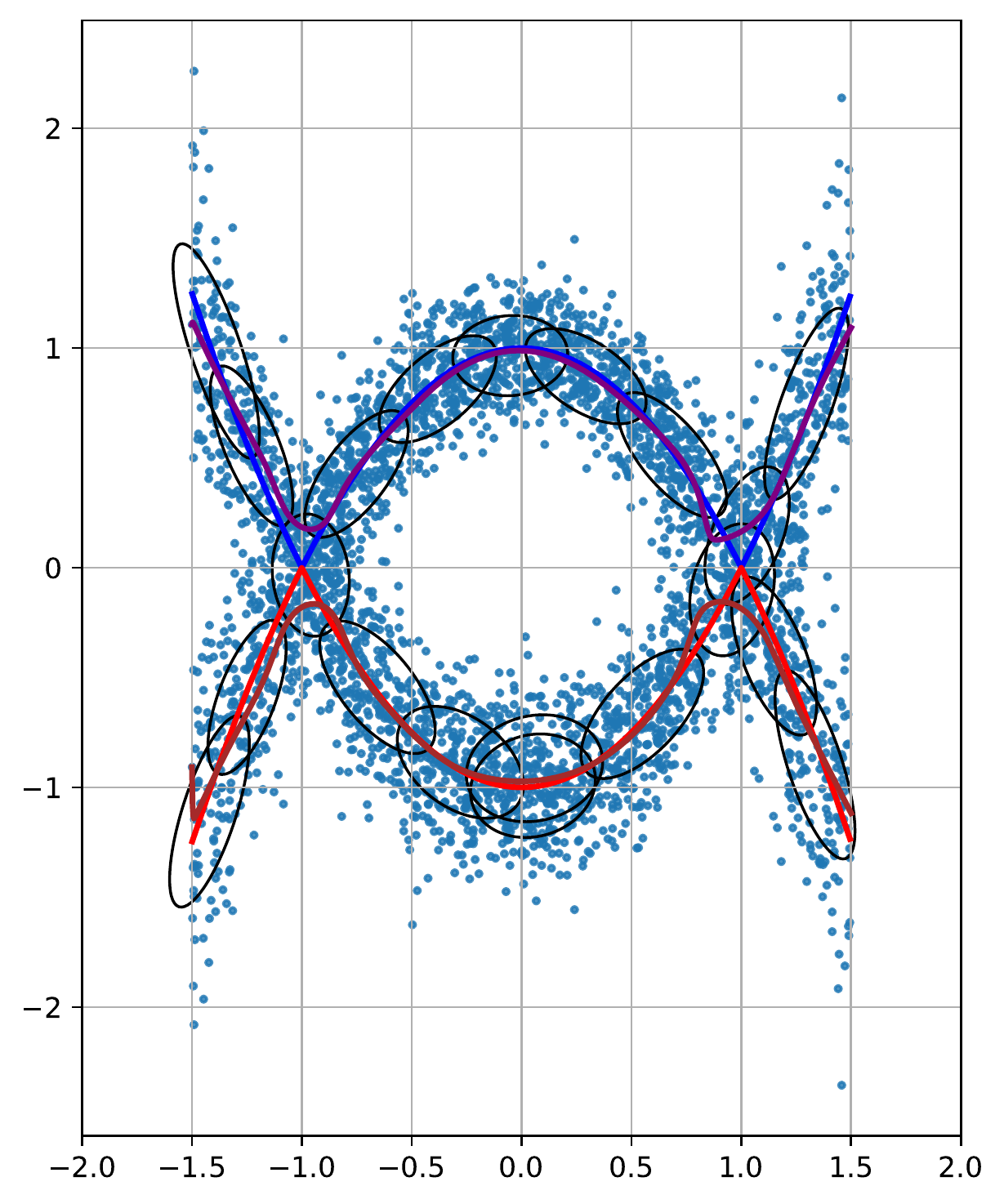}
  \caption{%
    Example of Gaussian Branching Regression (GBR). %
    We considered the algebraic variety $y^2 = (x-1)^2(x+1)^2$, which has symmetric positive (blue line) and negative (red line) branches. %
    We simulated 1000 noisy measurements (blue circles) with measurement noise $N(0,0.15)$ in both $x$ and $y$ with $x$ taken uniformly from $[-1.5, 1.5]$, and the $y$ branch chosen at random with equal probability. 
    Using \texttt{scikit.learn.gmm}, we represented the sample points as a GMM and plotted the covariance ellipses of the constituent Gaussians (black ellipses). %
    We then computed and plotted the predicted value of the model for a point traveling from $(-1,1)$ to the right, thus taking the positive (purple) branch, and similarly for a point traveling to the right from $(-1,-1)$ on the negative (dark red) branch. %
    \label{figXgmm}
  }
\end{figure}

Consider the algebraic variety $y^2 = (x - 1)^2 (x + 1)^2$ shown in Fig.~\ref{figXgmm}.
It consists of two branches: $y = (x - 1)(x + 1)$ and $y=-(x - 1)(x + 1)$, each of which is a manifold, and provides us with a toy example of a multi-valued function.
The variety itself as a point set in the $(x,y)$ plane is of measure 0, but a sampling process sampling $x$ uniformly in $[-1.5, 1.5]$, then taking a corresponding $y$ from either branch at equal probability, and finally corrupting both $x$ and $y$ with $N(0,0.15)$ Gaussian measurement noise, produces a measurable probility distribution on the $(x,y)$ plane. 
We built a GMM model of these data using \texttt{sklearn.mixture.GaussianMixture}, which produced a collection of Gaussian components $N(\mu_k, S_k)$ with their respective weights $w_k$.
We applied equation \ref{eqn:TLS} for each $S_k$ to obtain the slope $A_k$.
Note that $\mu_k$ is 2D, i.e. $\mu_k := ( \mu_{x,k}, \mu_{y,k})$.

Our GBR predicted value $\hat y(x, y_p)$ for the function at $x$ given the previous value of $y_p$ is:
\begin{align}
  c_k &:= w_k \Pr((x,y_p) | N(\mu_k, S_k)) \\
  \hat y (x, y_p) &:= \frac{1}{\sum_k c_k}\sum_k  c_k\left(  A_k (x - \mu_{x,k}) + \mu_{y,k} \right)
\end{align}

\subsection{Pre-processing to Produce a Ruled Surface}

As noted above, GBR is quite general.
It includes the ability to handle multi-valued functions, and because it is merely the application of the Nadaraya-Watson regressor to $f(x,y_p)$ it inherits the universality of this regressor.
The motility map, on the other hand, has some important additional structure that may be worth enforcing: its graph is a ruled surface since $v_b \ A(r)\dot r$ is linear in $\dot r$.

To enhance our algorithm's ability to produce the desired linear relationship from small samples, we first construct a GBR with the sample data, and then reprocess the data to create extrapolations of the linear relationship, as follows.
\begin{enumerate}
	\item Define $n_r$ to be the dimension of $r$, $n_v$ the dimension of $v_b$, and $n_{tot} := n_v + 2 n_r$ the total dimension of the space containing the graph. 
	Pick meta-parameters $\alpha>0$, $\beta>0$, and integer $c>0$
	\item Produce a GBR model $\{(w_k, \mu_k, S_k)\}_{k=1}^N$ consisting of $N$ Gaussians, where the $\mu_k$ is the mean of the $k$-th constituent Gaussian, whose variance is $S_k$ and whose relative weight is $w_k$. 
	For each of these Gaussians, compute $A_k$ such that $v_b = {\mu_{v_b}} + A_k (\dot r - \mu_{\dot r})$ the TLS estimate of the motility map $A(\mu_r)$.
	\item Define the Mahalanobis distance of the point $x = (v_b,r,\dot r)$ from Gaussian $k$ is $m_k(x) := (x-\mu_k)^\T S_k^{-1} (x-\mu_k) / n_{tot}$.
	A value of $m_k(x)<1$ implies the point is ``typical'' for that Gaussian, i.e. within one standard deviation from the mean.
	We will use $C_k := \{ x | m_k(x) \leq 1 \}$ as the points ``associated'' with the Gaussian $k$. 
	\item For each $C_k$ that contains $N_k$ data points, with $N_k \geq c $ do the following $\lceil \alpha N_k \rceil$ times to create an augmenting dataset ${\hat C}_k$:
	\begin{enumerate}
		\item Choose at random with replacement data points $(v_i,r_i,{\dot r}_i) \in C_k$, $i = 1\ldots c$
		\item Choose at random from ${\tilde b}_i \sim N(0,1)$ for $i = 1\ldots c$.
		  Define renormalized weights $b_i := {\tilde b}_i \left( \sum_{i=1}^c {\tilde b}^2_i \right)^{-1/2}$.
		  These correspond to a point uniformly sampled from the unit sphere in $c$ dimensions.
		\item Add the artificial point $\hat x := ({\hat v}_b,{\hat r},{\hat {\dot r}})$ to ${\hat C}_k$, given by:
			\begin{align}
				{\hat v}_b & := \mu_{k,v_b} + \beta \sum_{i=1}^c b_i (v_i-\mu_{k,v_b}) \nonumber\\
				{\hat r} & := \mu_{k,r} + \sum_{i=1}^c b_i (r_i-\mu_{k,r}) \nonumber\\
				{\hat {\dot r}} & := \mu_{k,\dot r} + \beta \sum_{i=1}^c b_i ({\dot r}_i-\mu_{k,\dot r})
			\end{align}
	\end{enumerate}
	\item Add the augmenting dataset to the original dataset and recompute the GBR to produce an ``Augmented GBR (A-GBR)'' 
\end{enumerate}

Assuming the Gaussians each have $c$ data points, by using this process we created an additional factor of $\alpha$ data points, each of which is the linear combination of $c$ points extrapolated by a factor of $\beta$.
If the original data was affine in its relationship between $\dot r$ and $v_b$, by augmenting it with a $\beta>1$ we extrapolated this relationship to a larger distance along the $\dot r$ directions.
By this we produced an A-GBR that respects this affine relationship over a larger range of $\dot r$ than a naive GBR model.

We needed to choose $\alpha$, $\beta$, and $c$.
We chose $c$ large enough to sufficiently reduce the noise of the affine model by averaging nearby points so that our extrapolation by factor $\beta$ seemed to provide a benefit.
We chose $\alpha$ arbitrarily, but small enough not to noticeably slow the computation time.
The parameters we used are: number of Gaussians $60$, $\alpha = 1$, $\beta = 1.2$, $c = 5$.

\begin{figure}[ht]
		\includegraphics[width=\columnwidth]{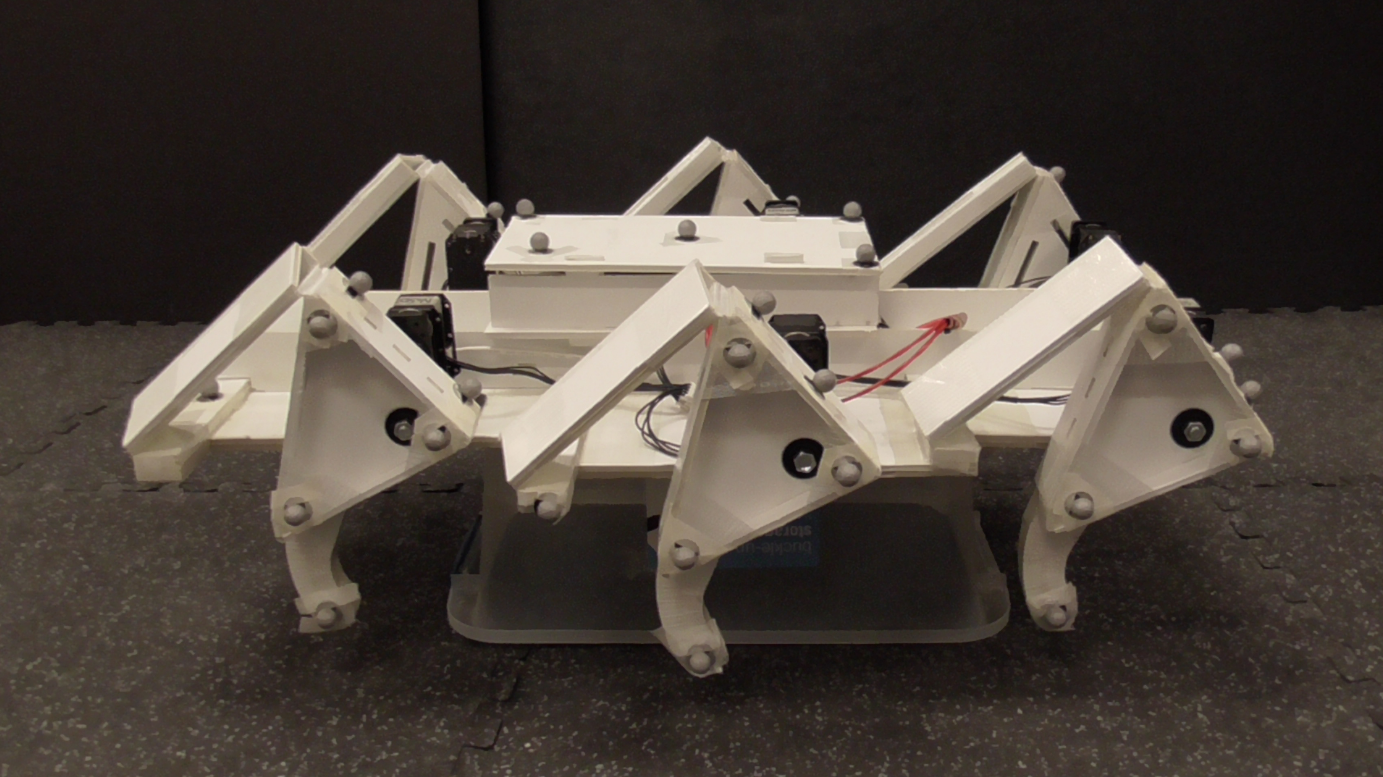}
		\caption{BigAnt Robot.%
		The BigAnt robot has six 1-DoF legs each using a 4-bar linkage which was carefully designed balancing leg clearance and the changes in gearing ratio \citep[ch. 2]{Zhao-2021-PhD}. %
		It was designed as a demonstration of the benefits of the ``Plates and Reinforced Flexures'' approach to robot design and fabrication \citep{Fitzner-2017-PARF}: its chassis uses only two materials (foam core board and fiber reinforced tape), costs about \$US 20, and takes a competent student about one day to make. %
		We have demonstrated the ability of this robot to go over rough terrain \href{https://youtu.be/RzQEB6N6pD8}{[Wave Field Video]} and through snow \href{https://youtu.be/RDDGD3j_uCs}{[BigAnt in Snow Video]}.
		}
		\label{fig:bigant}
\end{figure}

\begin{figure}[ht]
		\includegraphics[width=\columnwidth]{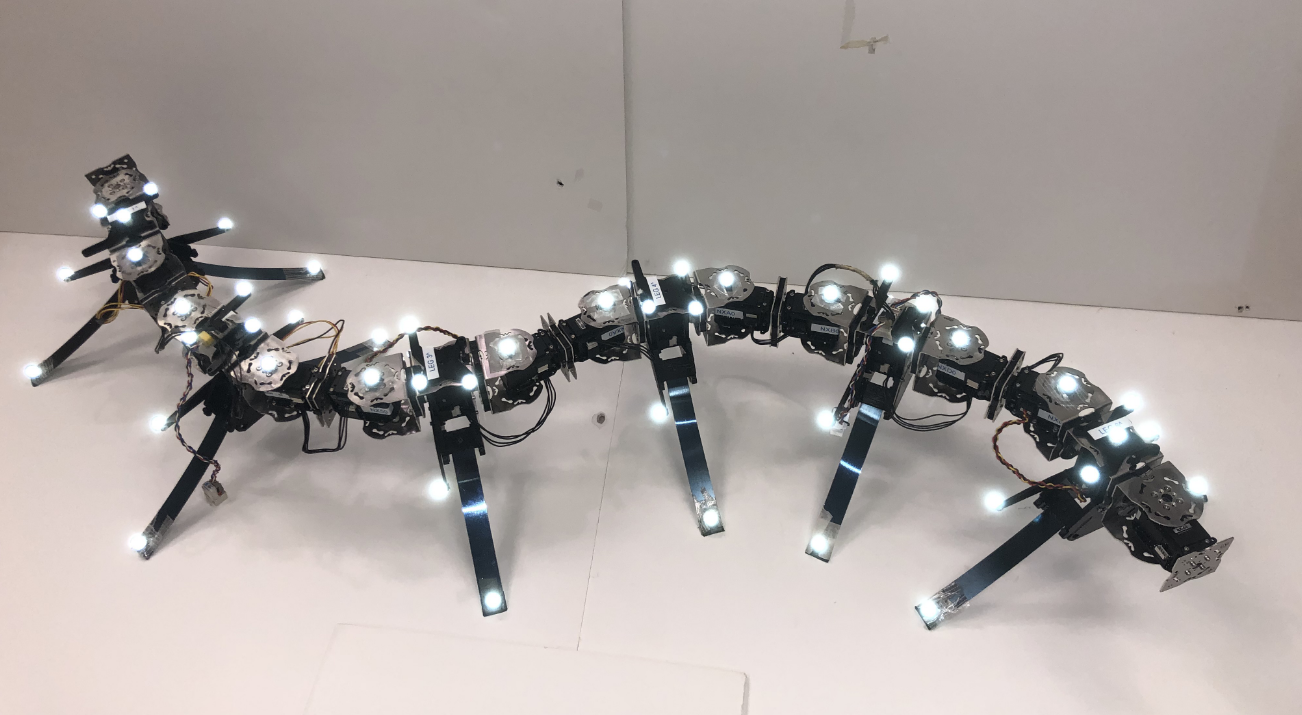}
		\caption{Multipod Robot. %
		  The Multipod robots are a collection of segmented modular robots with spring steel legs, going from 4 legs (2 pairs) through to 12 legs (6 pairs). %
			These robots were used to study how locmotion dynamics changes when the number of legs increase, leading in part to the results of \citet{zhao2022walking}.
		}
		\label{fig:Multipod}
\end{figure}

\begin{figure*}[h]
	\centering
	\includegraphics[height=7cm, trim={0 0 3.3cm 0},clip]{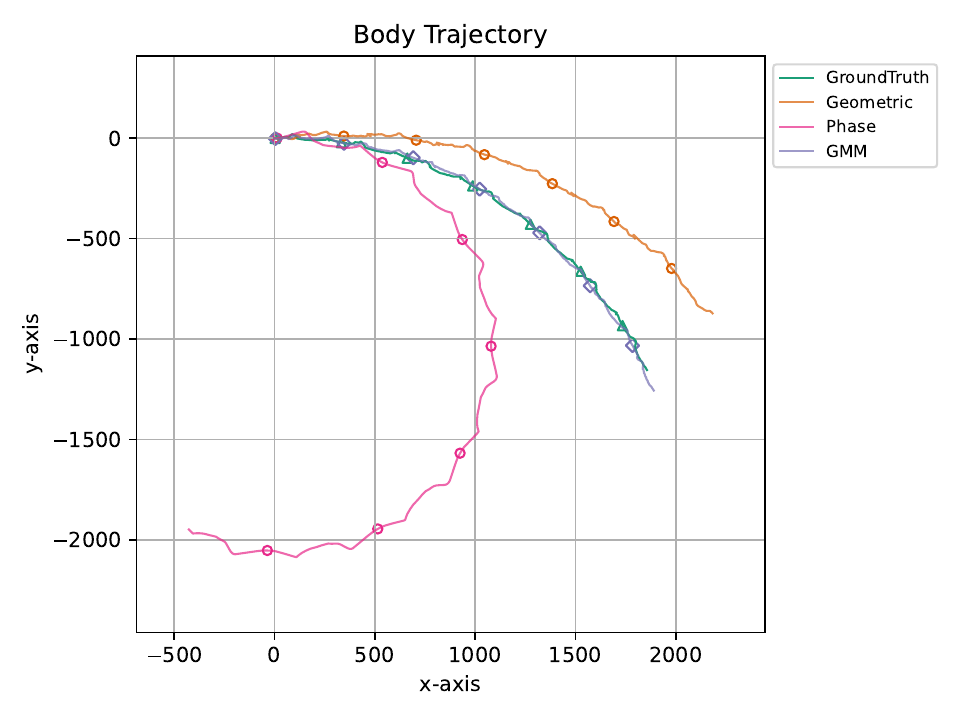}
	\includegraphics[height=7cm]{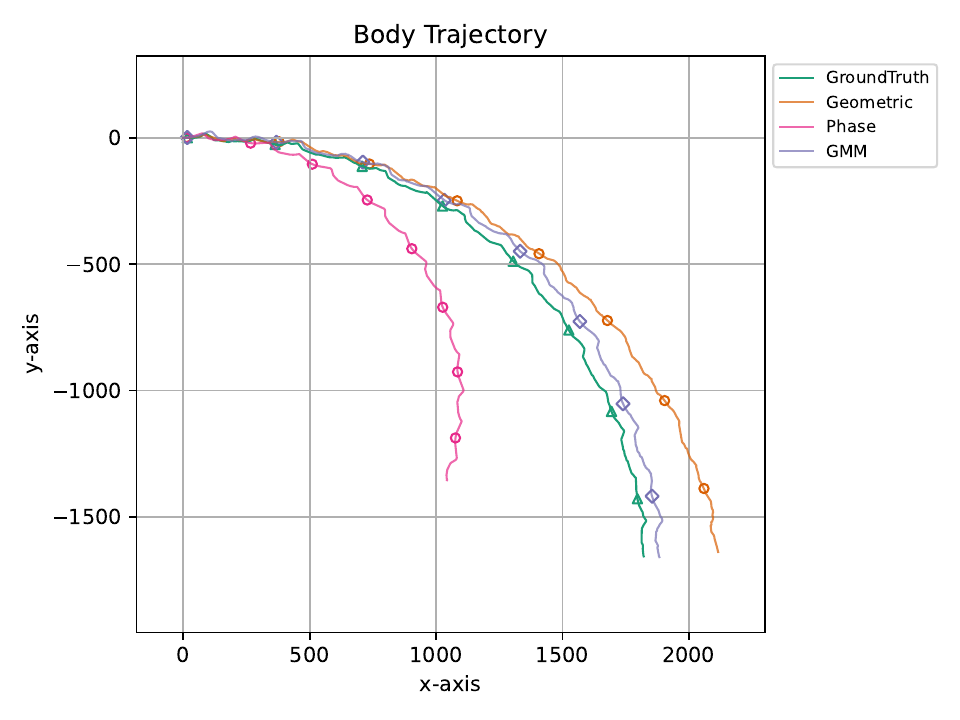}
	\includegraphics[height=7cm, trim={0 0 3.3cm 0},clip]{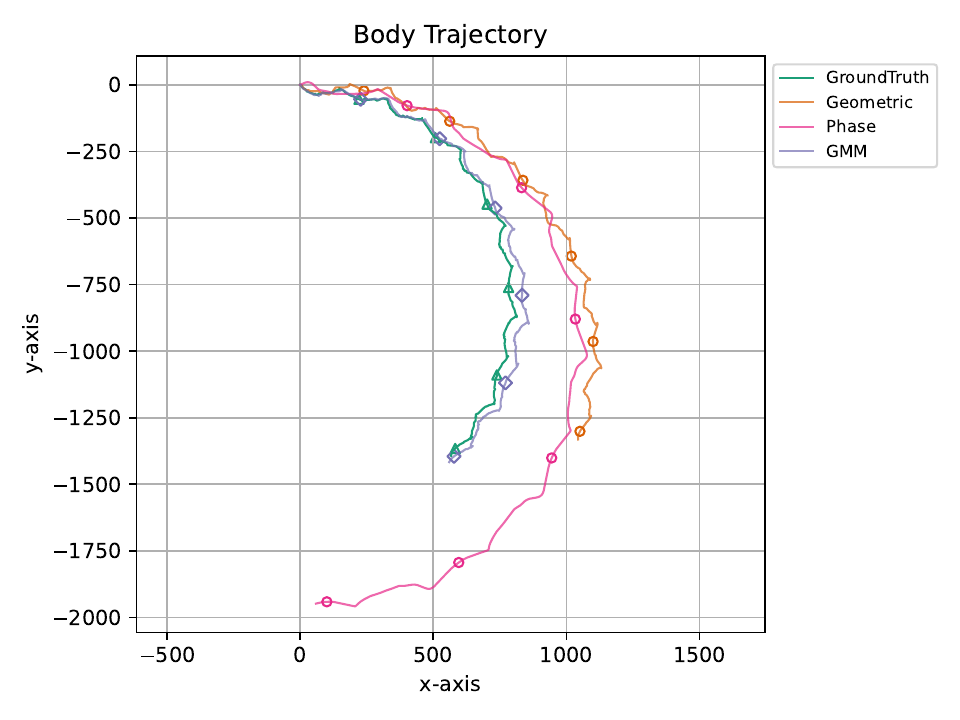}
	\includegraphics[height=7cm]{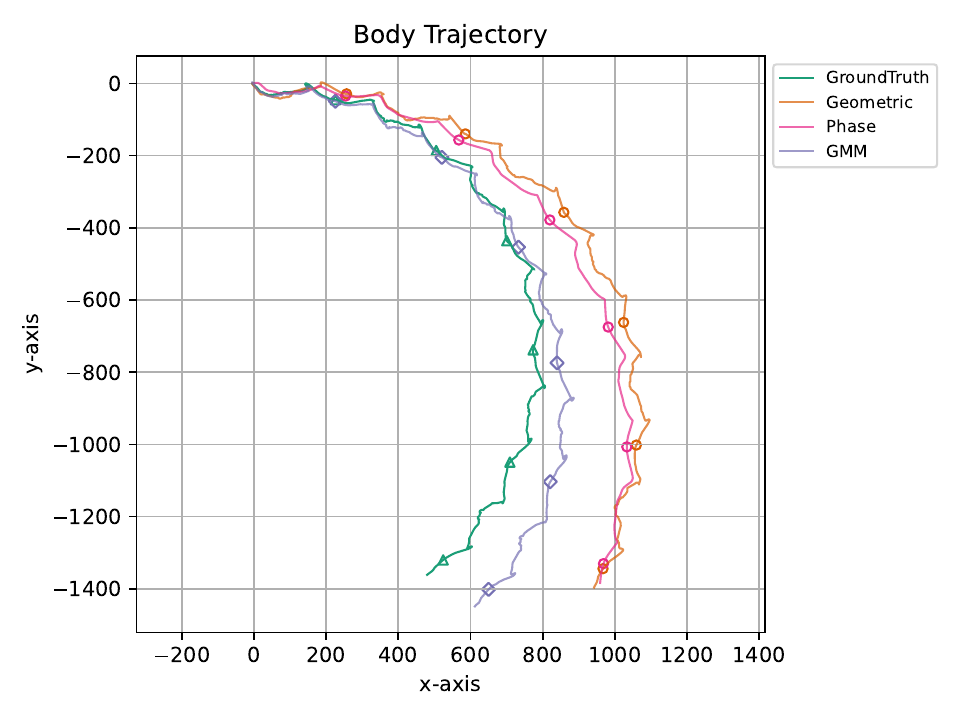}
	\caption{%
		Comparison of model extrapolation to less and more turning. %
		We compared the predictions of the models for a recording of $6$ strides (cycles) of motion, after training on data from an intermediate turning rate (steering parameter $s=0.50$ of \cite{zhao2020multi}). %
		We attempted to predict both noticeably lesser ($s=0.25$; top) and noticeably greater ($s=0.75$; bottom) turning rates;. %
		In each case we plotted the ground truth motion capture body centroid (thick green lines, triangle markers), the Geometric model (orange lines, circle markers), the Phase model (pink lines, circle markers), and the A-GBR model (purple lines, diamond markers)  %
		We plotted the markers at the same phase in each cycle of motion, to make it easier to understand how much the robot moved in a cycle. %
		Results demonstrate that while the Geometric model is noticeably better than the Phase model, A-GBR is noticeably superior to both.
		\label{fig:bigAntTraj}
	}
\end{figure*}

\begin{figure*}[h]
	\centering
	\includegraphics[height=7cm, trim={0 0 3.3cm 0},clip]{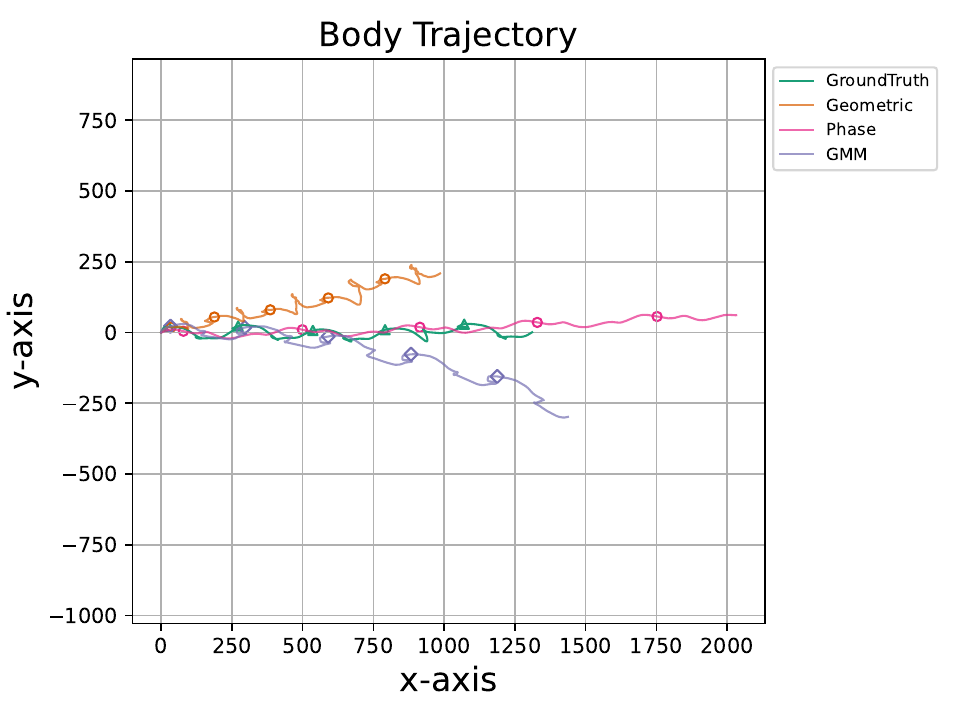}
	\includegraphics[height=7cm]{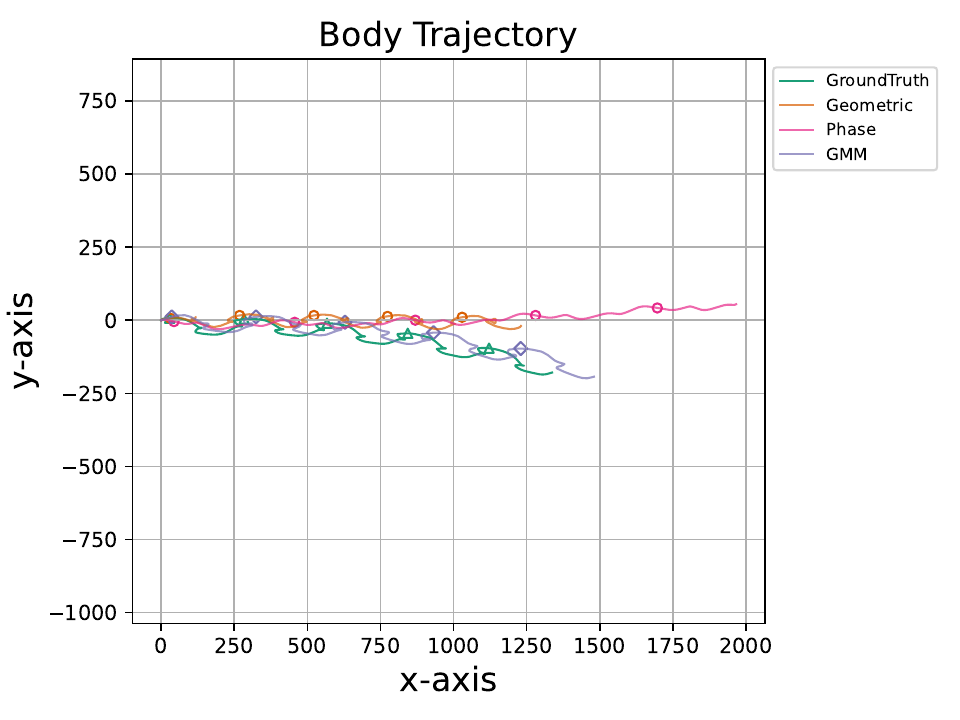}
	\includegraphics[height=7cm, trim={0 0 3.3cm 0},clip]{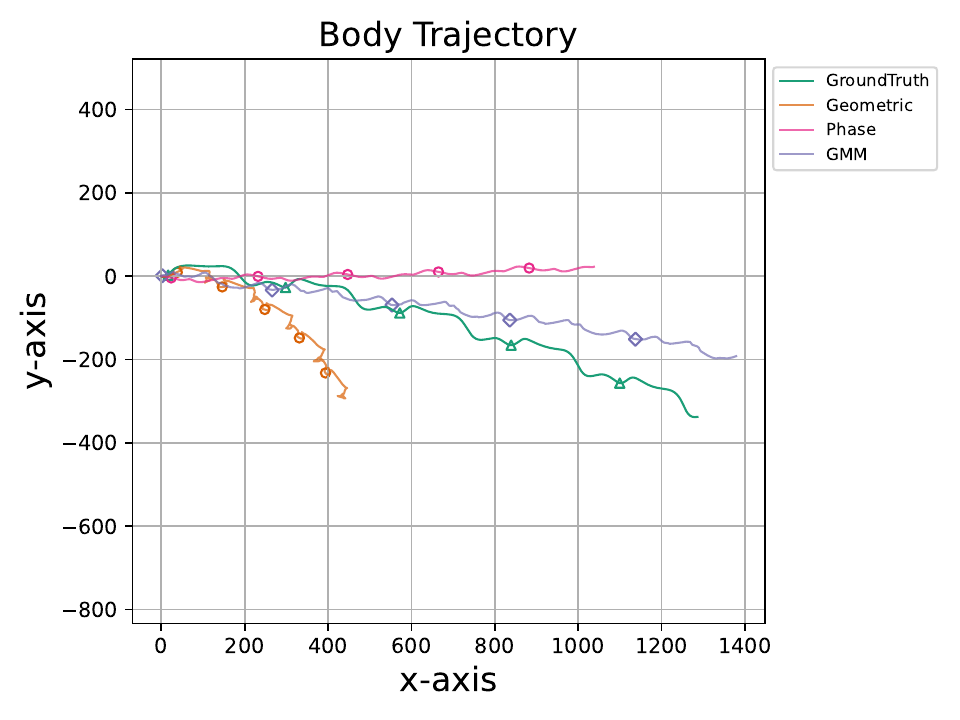}
	\includegraphics[height=7cm]{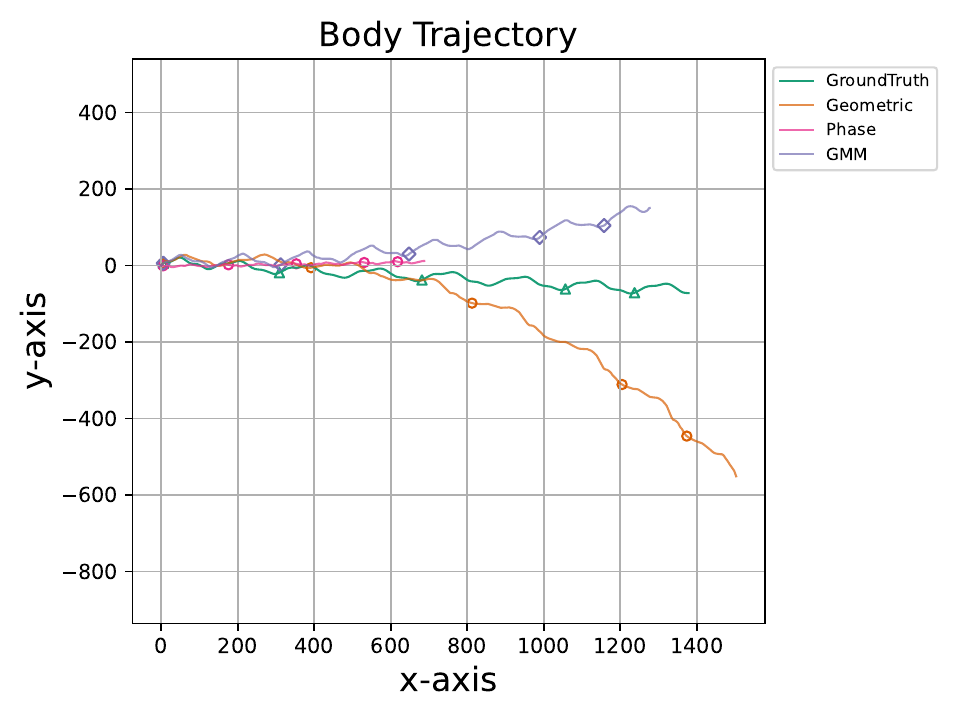}
	\caption{%
		A-GBR performance on the Multipod. %
		We trained A-GBR with data from intermediate phase offset gaits ($\Delta \phi = [1.05\pi, 1.1\pi, 1.15\pi]$) and tested its performance on gaits with both lower and higher $\Delta\phi$. %
		We provide similar trajectory plot as in Fig.~\ref{fig:bigAntTraj} (top plots $\Delta \phi = [0.85\pi, 0.9\pi, 0.95\pi, 0.97\pi, \pi]$; bottom plots $\Delta \phi = [1.2\pi, 1.25\pi, 1.3\pi, 1.35\pi, 1.4\pi]$). 
		While the Geometric model is noticeably better than the Phase model, A-GBR is noticeably superior to both.
	}
	\label{fig:MultipodTraj}
\end{figure*}

\subsection{Model types}

Since our goal is to improve the ability to construct data-driven motility map models, we chose to compare robot motion tracking data (``ground truth'') to our A-GBR estimates, and compared these estimates to estimates obtained from the approach of \cite{bittner2018geometrically} and \cite{zhao2022walking}.
We challenged our models with extrapolation, similar to \cite{zhao2022walking}, by building a model using training data from an intermediate turning gait, and testing against both sharper and more moderate turns.
The ability to extrapolate well is critical for making a model usable for the optimization framework shown in \cite{bittner2018geometrically} and \cite{bittner2021optimizing}.

\subsubsection{Geometric Model}
For each training dataset, we computed a motility map model following the construction in \cite{bittner2018geometrically}, namely first separating the data into bins by ``phase'', i.e. where the robot was within its cycle of motion.
For each bin we computed the average body velocity, and built a first order Taylor series expansion of the motility map around this phase-averaged behavior.
Finally, we interpolated the bin averages and the tensors representing the Taylor series using a Fourier series to produce a body velocity estimation function of the form $v_b(\phi,r,\dot r) := v_0(\phi) + \Delta(\phi,r) \dot r$ giving body velocity as a function of phase, shape, and shape velocity.
Here the average body velocity as a function of phase was $v_0(\phi)$, and we added a first order correction $\Delta(\phi,\cdot)\cdot$.

\subsubsection{Phase Model}
The ``Phase'' model uses only the the phase-averaged $v_b$, i.e. is equivalent to the ``Geometric'' model with $\Delta := 0$.

\subsection{Loss Function}

To select a good set of meta-parameters, we chose a loss function representing a non-dimensionalized modeling error.
For each stride, i.e. full gait cycle, we consider the forward (initial body frame $x$), sideways (initial body frame $y$), and heading (initial body frame $\theta$) motion produced.
Ignoring any possible correlation of these variables, we took the loss to be the total $z$-score of the $x$, $y$, and $\theta$ predictions with respect to the corresponding variable in the ground truth data.
This means that with our choice of loss function, the loss of a random permutation of the ground truth data against itself was set to approach $3$, and the loss of a perfect prediction is zero.

\section{Results}

To evaluate the performance of our model, we present results using motion capture data from the BigANT (Fig. \ref{fig:bigant}) \cite{BigAntData} and Multipod (Fig. \ref{fig:Multipod}) \cite{MultipodData} robots. 

Data for both BigANT and Multipod were collected using a Qualisys motion capture system consisting of 10 Oqus-310+ cameras operating at 120~fps as described in \cite{BigAntData, MultipodData}. 
We trained the Phase, Geometric, and A-GBR models using shape space data represented by motion tracking marker locations relative to a body frame.
The body frame we used in each case was the body frame available in the publicly available dataset; optimal selection of the body frame, as in
\cite{hatton2011geometric}, might improve on our results.

\subsection{BigANT Results}

BigANT is a hexapod robot with one degree of freedom per leg.
The analyzed gaits induce BigANT to steer in varying rates, as defined by \cite{zhao2020multi}.
In this study, we assess the extrapolation capability of A-GBR by training it on data with an intermediate steering rate and testing it on both lower and higher steering rate recordings.

\begin{figure}[H]
	\centering
	\includegraphics[width=.43\textwidth]{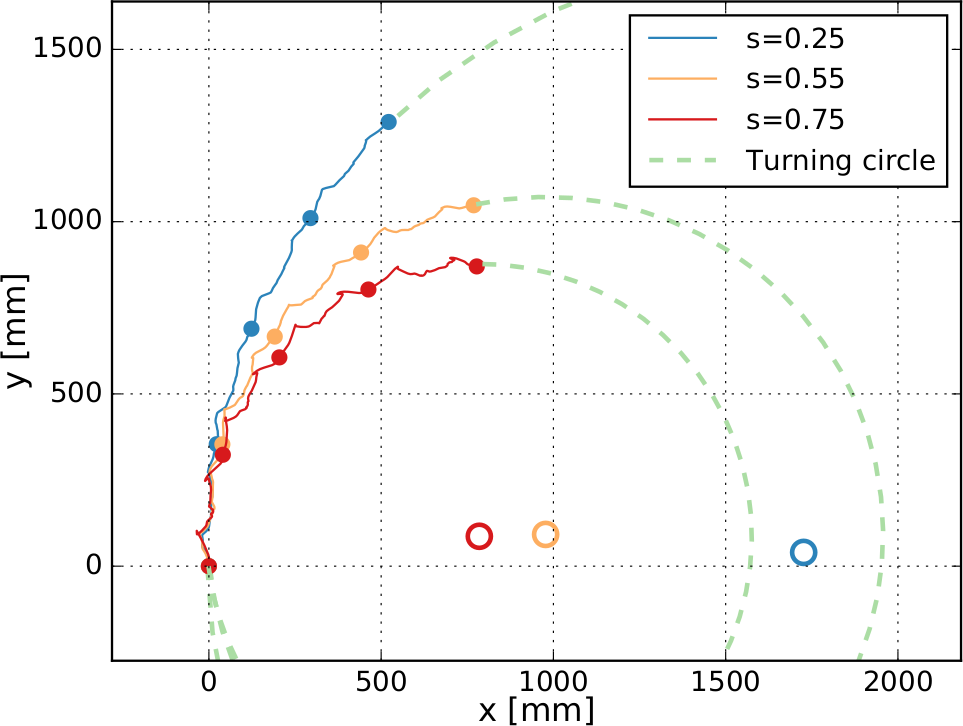}
	\caption{Turning rate demonstration.
		The different turning radii are demonstrated by a circle fitted to the motion tracking data (dashed green line and circular markers for centers; right plot; reproduced from Fig. 3.9 of \cite{Zhao-2021-PhD}).
	}
	\label{figXtraj}
\end{figure}

To get a general sense of the difference between the three model types when applied to BigANT, we provided Fig. \ref{figXtraj} and Fig. \ref{fig:bigAntTraj} which shows examples of observed body centroid trajectories and the predicted motion based on the various models.

\subsection{Multipod Results}

The Multipod robots are a family of serpentine multi-legged robots varying from 4 to 12 legs. 
Each leg is (functionally) movable in 2 DoF, and the robots move by commanding the legs in a traveling wave defined by the phase offset between adjacent segments. 
In this study, we focus on the YRY (yaw-roll-yaw) configuration 6-legged Multipod, as defined by \cite{zhao2020multi, Zhao-2021-PhD}. 
We trained A-GBR on gaits intermediate phase offsets ($\Delta \phi = [1.05 \pi, 1.1 \pi, 1.15\pi]$) and tested on both lower ($\Delta \phi = [0.85 \pi, 0.9 \pi, 0.95\pi, 0.97\pi, \pi]$) and higher ($\Delta \phi = [1.2 \pi, 1.25 \pi, 1.3\pi, 1.35\pi, 1.4\pi]$) phase offsets, as defined by \cite{Zhao-2021-PhD}. 
Figure \ref{fig:MultipodTraj} illustrates the robot body frame trajectories corresponding to these different gaits.

\subsection{Prediction and Residual Error Distributions}
\begin{figure*}[h]
  \centering
  \includegraphics[width=.45\textwidth]{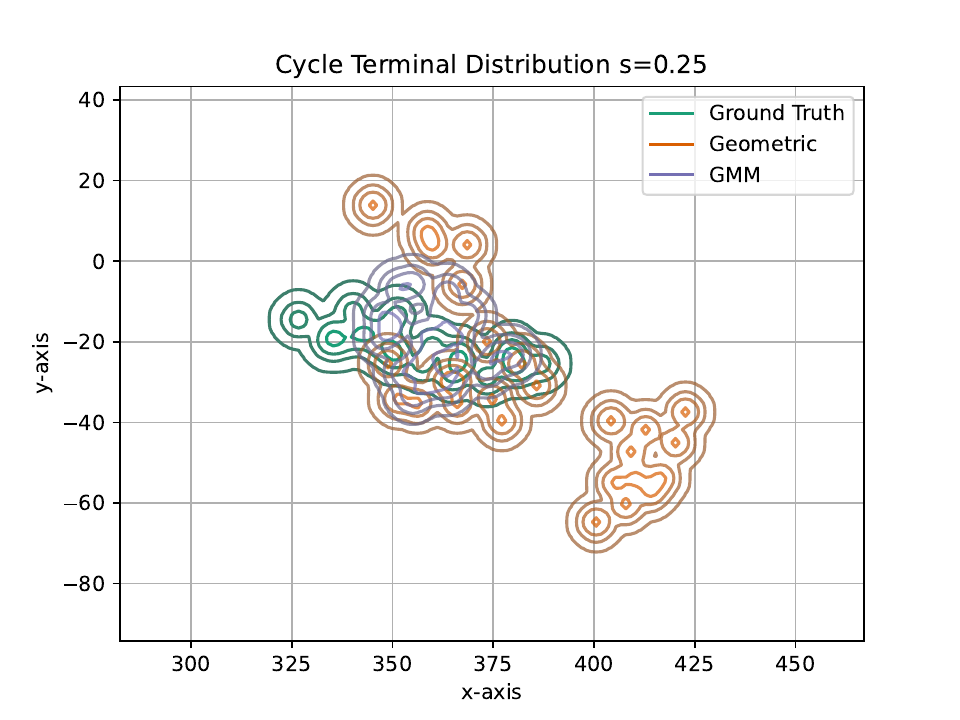}
  \hspace{3mm}
  \includegraphics[width=.45\textwidth]{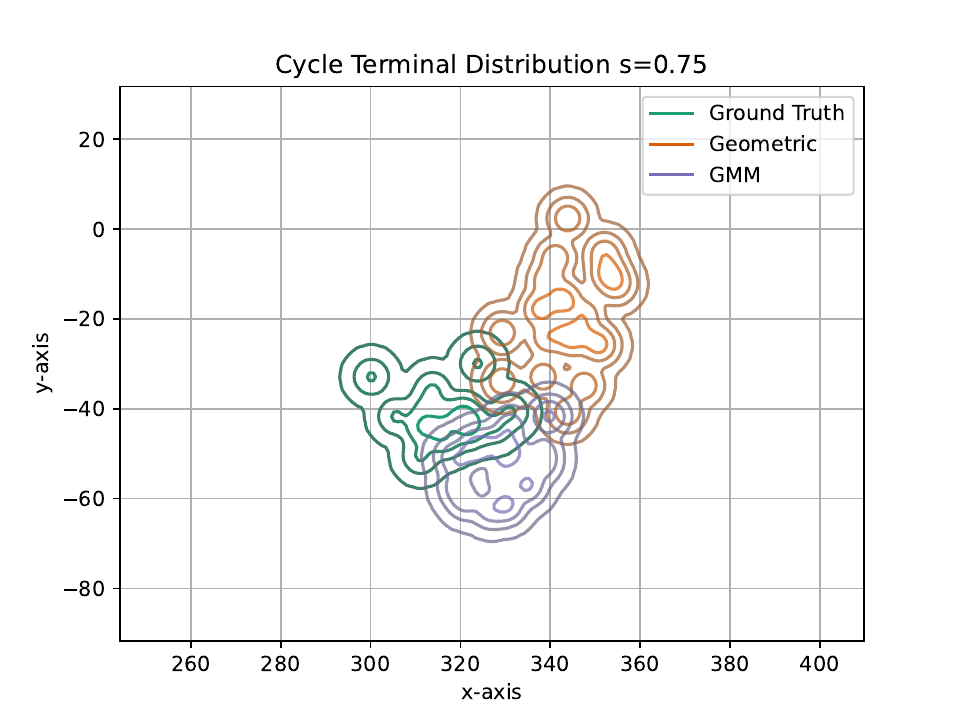}\\
  \includegraphics[width=.45\textwidth]{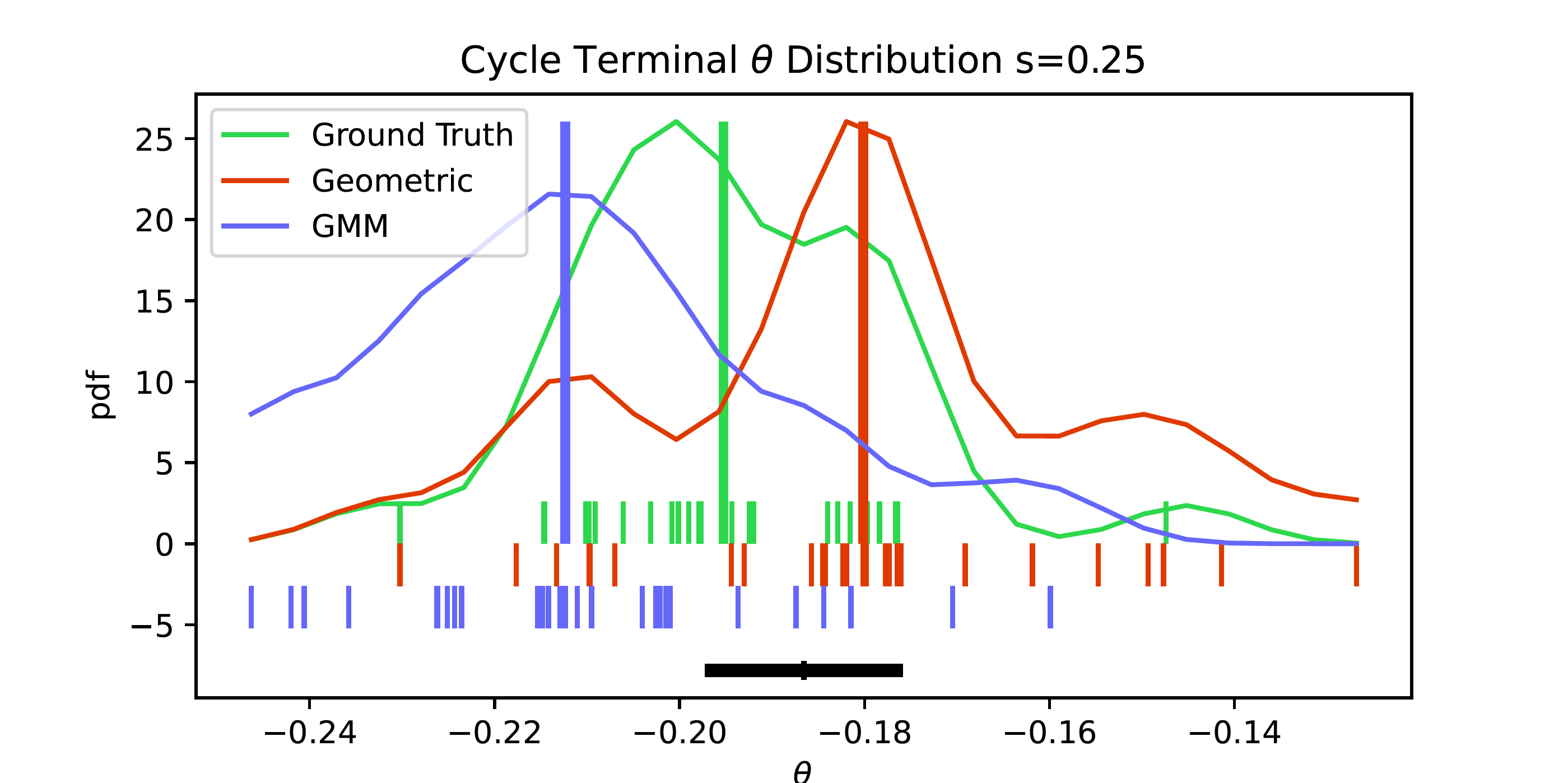}
  \hspace{3mm}
  \includegraphics[width=.45\textwidth]{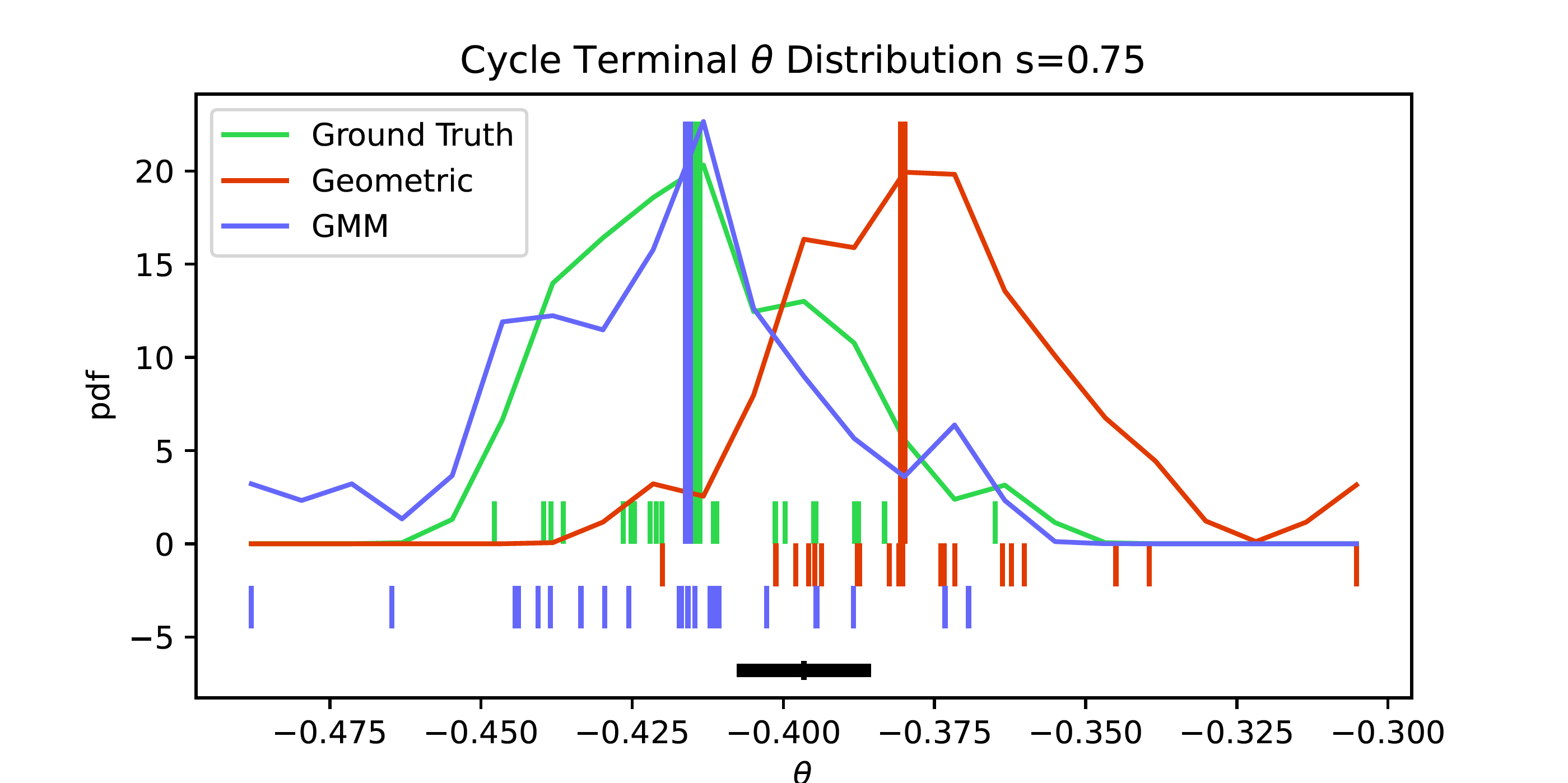}
  \caption{Distribution of final positions after a cycle of BigANT motion. %
		We ploted a kernel-smoothed density estimate (KSDE) of the forward ($x$), sideways ($y$), and turn ($\theta$) motion produced by each stride (i.e. full gait cycle). %
  	We plotted ground truth (green), the \cite{bittner2018geometrically} gait-based geometric model prediction (orange), and the GMM model prediction (purple). %
		We trained the models on intermediate turning data (``turn parameter'' $s=0.50$) and plotted the extrapolated results for a slower turn ($s=0.25$; $x$, $y$, $\theta$ left) and a faster turn ($s=0.75$; $x$, $y$, $\theta$ right). %
  	For the $\theta$ distribution we plotted the observed values (vertical tics at the bottom of the plots), the distribution medians (thick vertical lines), and a scale bar for the width of the smoothing kernel (thick black line).\label{fig:bigAntTerminal}
  }
\end{figure*}

\begin{figure*}[h]
  \centering
  \includegraphics[width=.45\textwidth]{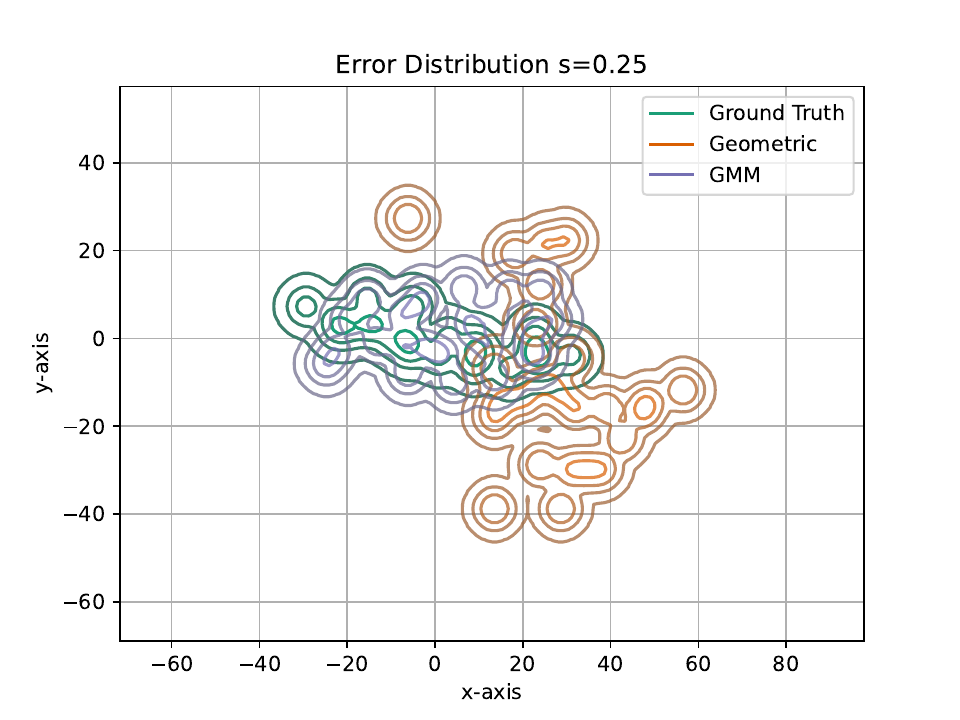}
  \hspace{3mm}
  \includegraphics[width=.45\textwidth]{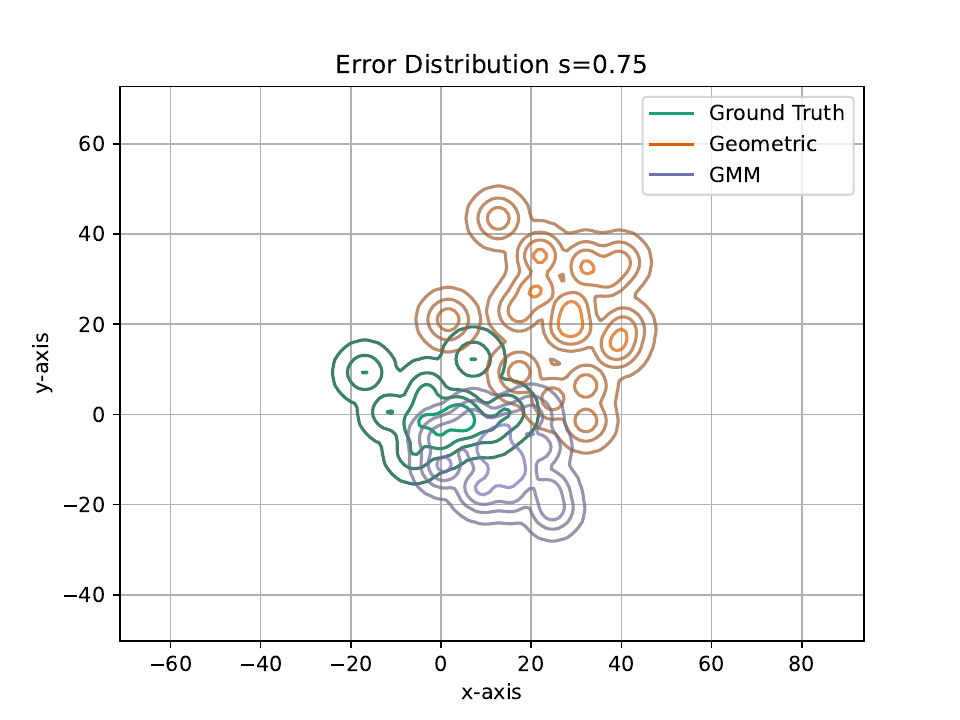}\\
  \includegraphics[width=.45\textwidth]{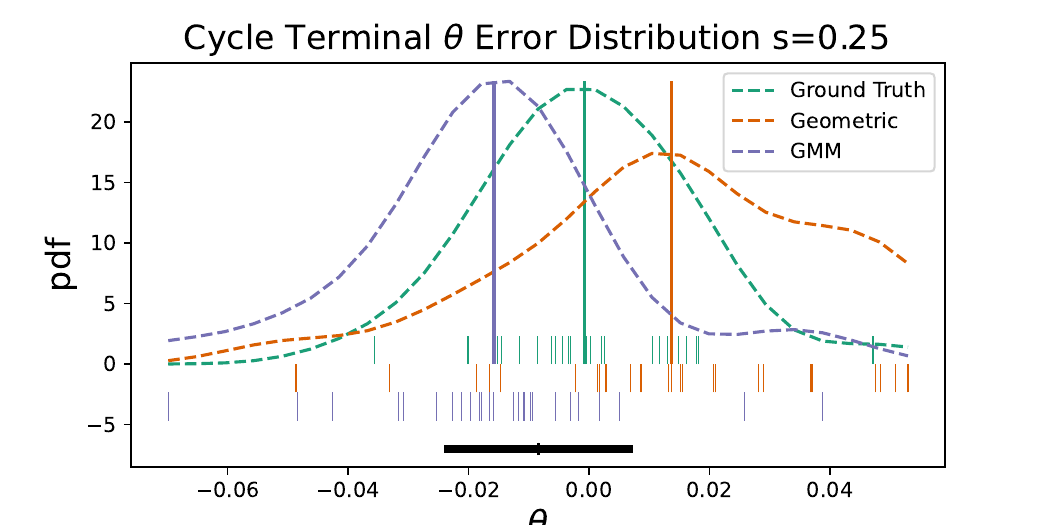}
  \hspace{3mm}
  \includegraphics[width=.45\textwidth]{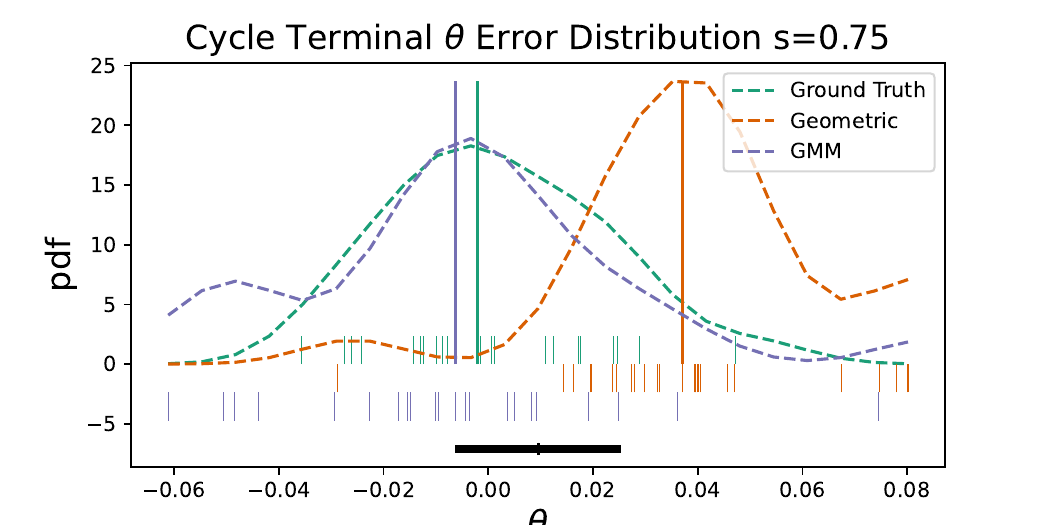}
  \caption{Distribution of the prediction residual of final positions after a cycle of BigANT motion. %
		We ploted a KSDE of residual error distributions of the forward ($x$), sideways ($y$), and turn ($\theta$) motion produced by each stride (i.e. full gait cycle). %
		To provide a reference variance allowing the relative size of the modeling error to be understood by the reader, we plotted the ground truth data minus its mean in the error distributions. %
		The model and training setting are the same as in Fig.\ref{fig:bigAntTerminal}. %
  		For the $\theta$ distribution we also plotted the observed values, the distribution medians, and a scale bar for the width of the smoothing kernel as in Fig.\ref{fig:bigAntTerminal}. \label{fig:bigAntError}
  }
\end{figure*}

\begin{figure*}[h]
  \centering
  \includegraphics[width=.45\textwidth]{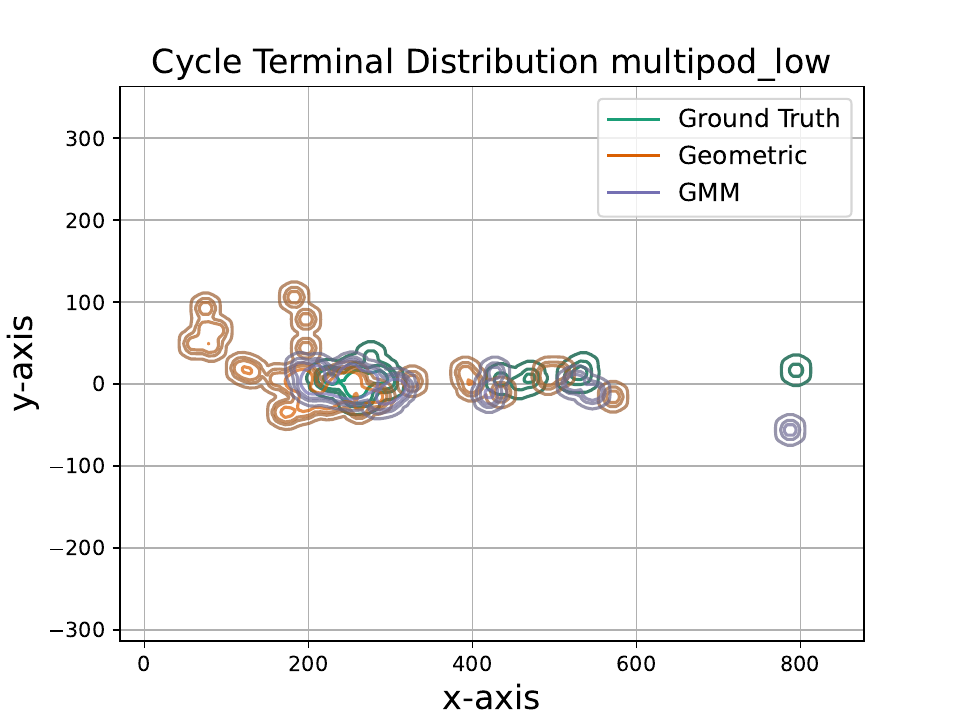}
  \hspace{3mm}
  \includegraphics[width=.45\textwidth]{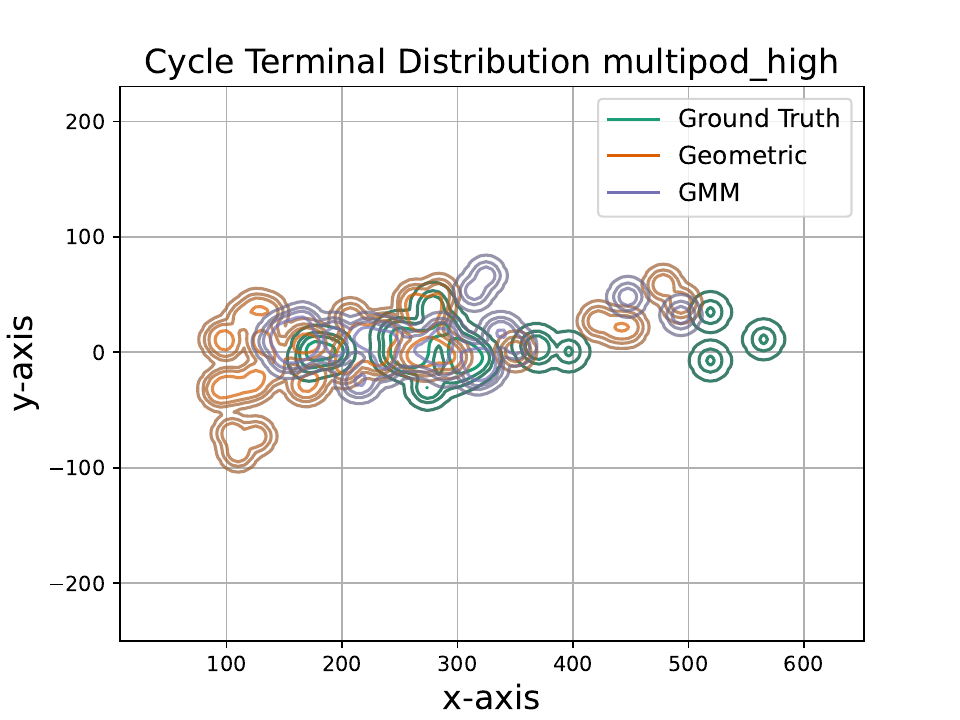} \\
  \includegraphics[width=.45\textwidth]{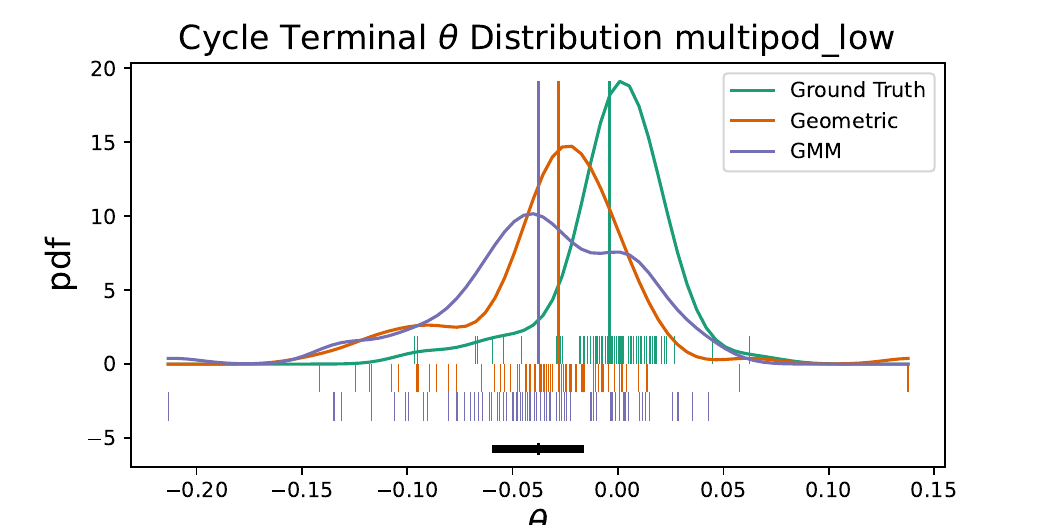}
  \hspace{3mm}
  \includegraphics[width=.45\textwidth]{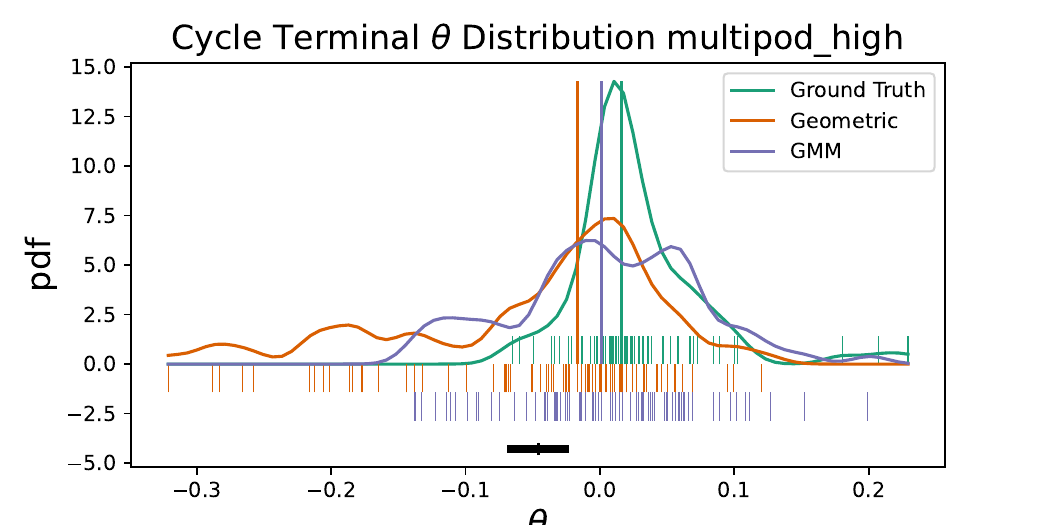}
  \caption{Distribution of final positions after a cycle of Multipod motions. %
		We plot a KSDE of the forward ($x$), sideways ($y$), and turn ($\theta$) motion produced by each stride (i.e. full gait cycle). %
		The model setting is the same as in Fig. \ref{fig:bigAntTerminal}.
		We trained the models on intermediate phase offset ($\Delta \phi = [1.05\pi, 1.1\pi, 1.15\pi]$) and plotted the extrapolated results for lower phase offsets ($\Delta \phi = [0.85\pi, 0.9\pi, 0.95\pi, 0.97\pi, \pi]$, left) and higher phase offsets ($\Delta \phi = [1.2\pi, 1.25\pi, 1.3\pi, 1.35\pi, 1.4\pi]$, right). %
		For the $\theta$ distribution, we present both the predicted values and the corresponding residuals.
  }
  \label{fig:MultipodTerminal}
\end{figure*}

\begin{figure*}[h]
  \centering
  \includegraphics[width=.45\textwidth]{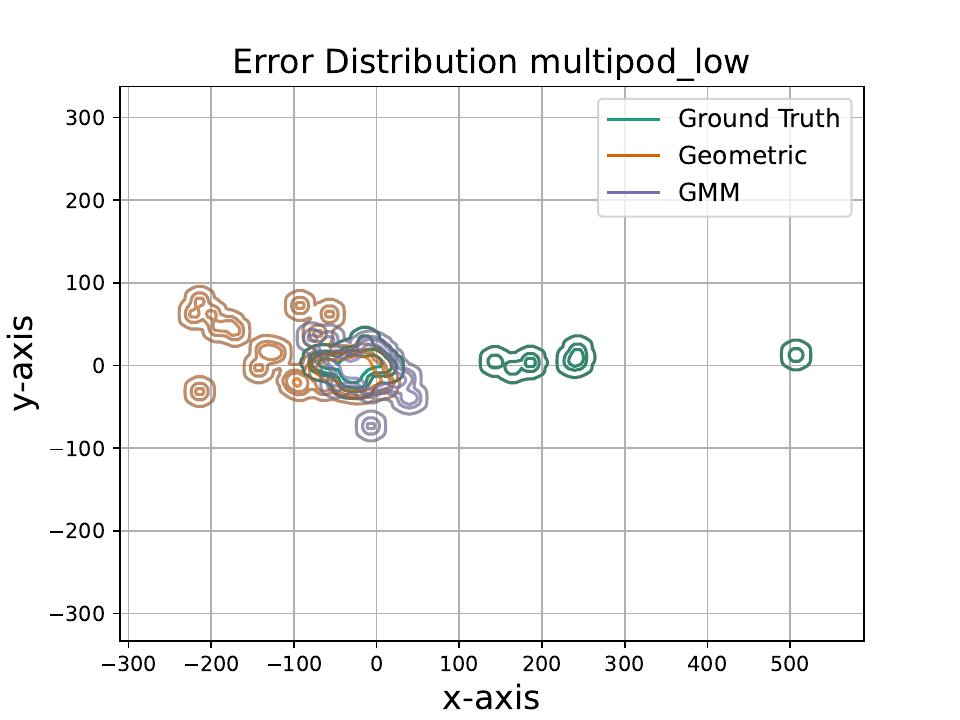}
	\hspace{3mm}
  \includegraphics[width=.45\textwidth]{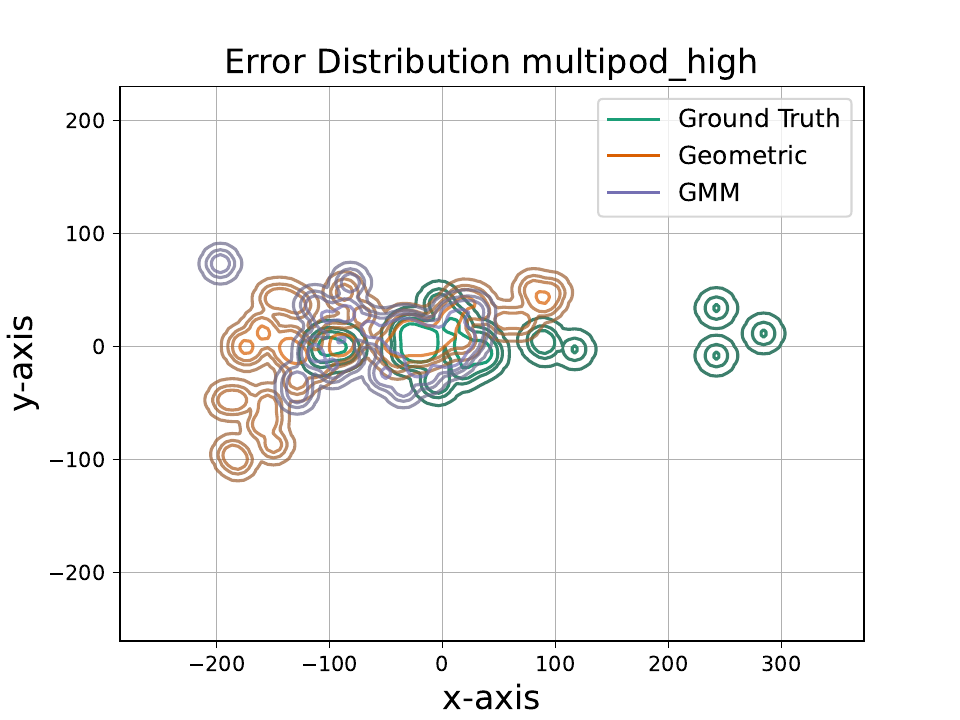}\\
  \includegraphics[width=.45\textwidth]{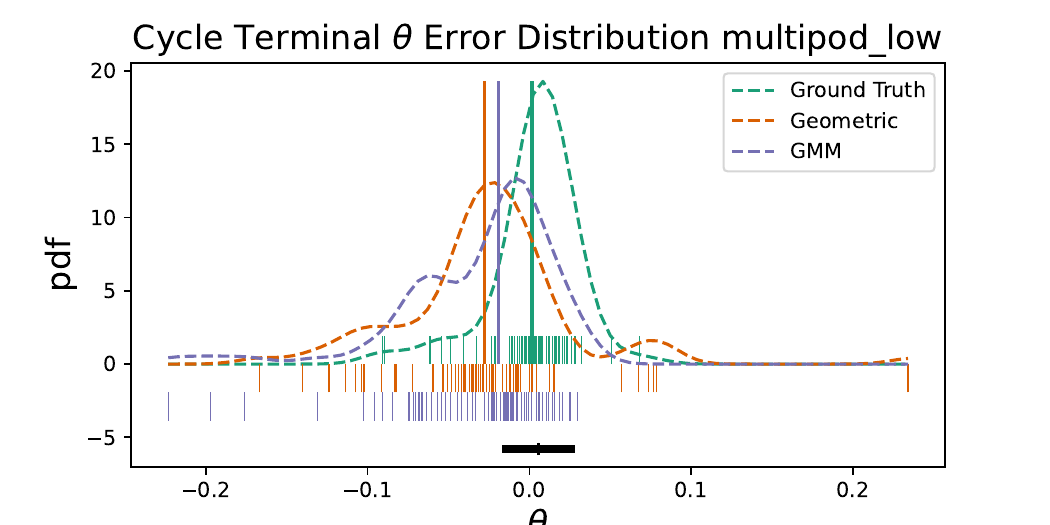}
  \hspace{3mm}
  \includegraphics[width=.45\textwidth]{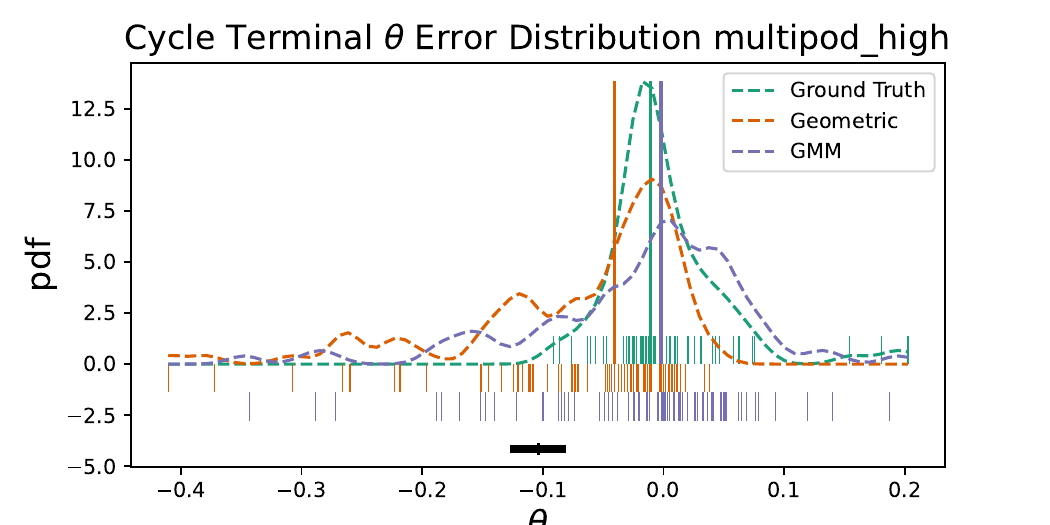}
  \caption{Distribution of the prediction residual of final positions after a cycle of Multipod motions. %
		We plot a KSDE of of residual error distributions of the forward ($x$), sideways ($y$), and turn ($\theta$) motion produced by each stride (i.e. full gait cycle) %
		The model and training setting are the same as in Fig. \ref{fig:MultipodTerminal}.
		For the $\theta$ distribution, we present 1-D KSDE of the residuals as described in Fig. \ref{fig:bigAntError}.
  }
  \label{fig:MultipodError}
\end{figure*}

We obtained better insight into the predictive ability of these models (BigANT Fig.~\ref{fig:bigAntTerminal} and Fig.~\ref{fig:bigAntError}; Multipod Fig.~\ref{fig:MultipodTerminal} and Fig.~\ref{fig:MultipodError}) by considering the distributions of the predictions of each model and the distributions of the residual errors of the models plotted on the backdrop of the distribution of the ground truth against its mean for scale.
We plotted these as kernel-smoothed density plots to give a (non-parametric) sense of how well these predicted distributions overlap.

\section{Discussion}

Our results show that using A-GBR to model the motility map produced predictions that were substantially more accurate than the previous data-driven geometric mechanics modeling approaches cited.
We have shown this for an \emph{extrapolation} problem --- for motions substantially different from the training data (see Fig.~\ref{fig:bigAntTraj}, Fig.~\ref{fig:MultipodTraj} and Fig.~\ref{figXtraj}), suggesting that these models can be useful for planning and for iterative gait optimization similar to that shown in \cite{bittner2018geometrically, bittner2021optimizing}.

The method we used here has significant advantages beyond the improved performance: 
[1] it does not rely on phase estimation or on having nearly periodic data as an input, allowing any motion data available to be incorporated;
[2] it can handle hysteresis effects in its input data;
[3] it provides meta-parameters to control how much linear extrapolation it applies to the parts of the motility map that are expected to be linear.

Obviously, the results here are preliminary and much work remains.
Our modeling approach needs to be tested with a variety of datasets, including testing with data that contain notable hysteresis effects, and data that are less linear in the $\dot r$ dependence of the motility map.
It would be natural to extend the models to the Shape Underactuated Dissipative Systems of \cite{bittner2022data}, and use methods for online update of GMM-s to produce an online adaptive version of A-GBR.
Another natural direction is to automate and improve the selection of meta-parameters, and to try other established manifold learning approaches for constructing models.

\subsection*{Funding}
Funding for this work was provided by NSF CPS 2038432; dataset was collected with funding from ARO MURI W911NF-17-1-0306 using equipment funded by ARO DURIP W911NF-17-1-0243.

\bibliographystyle{plainnat}
\bibliography{ref.bib}

\end{document}